**Artificial Intelligence–Enabled Analysis of Radiology Reports: Epidemiology and Consequences of Incidental Thyroid Findings**

Felipe Larios[1], MD*; Mariana Borras-Osorio[1], MD*; Yuqi Wu[2], PhD; Ana Gabriela Claros[1], MD; David Toro-Tobon[1, 3], MD; Esteban Cabezas[1], MD; Ricardo Loor-Torres[1], MD; Maria Mateo Chavez[1], MD; Kerly Guevara Maldonado[1], MD; Luis Vilatuna Andrango[1], MD; Maria Lizarazo Jimenez[1], MD; Ivan Mateo Alzamora[1], MD; Misk Al Zahidy[1], MS; Marcelo Montero[1], MD; Ana Cristina Proano[1], MD; Cristian Soto Jacome[1, 4], MD; Jungwei W. Fan[2], PhD; Oscar J. Ponce-Ponte[1, 5], MD; Megan E. Branda[6], MS; Naykky Singh Ospina[7], MD, MS; Juan P. Brito[1, 3], MD, MS.

*These authors contributed equally to the work.
Affiliations:
1. Care and AI Laboratory, Knowledge and Evaluation Research Unit, Division of Endocrinology, Diabetes, Metabolism, and Nutrition, Department of Medicine, Mayo Clinic, Rochester, Minnesota
2. Department of Artificial Intelligence and Informatics, Mayo Clinic, Rochester, Minnesota
3. Division of Endocrinology, Diabetes, Metabolism, and Nutrition, Mayo Clinic, Rochester, Minnesota
4. Department of Medicine, Norwalk Hospital, Norwalk, Connecticut
5. Derriford Hospital, University Hospitals Plymouth NHS Trust, Plymouth, United Kingdom
6. Division of Biomedical Statistics and Informatics, Department of Health Sciences Research, Mayo Clinic, Rochester, Minnesota
7. Division of Endocrinology, Department of Medicine, University of Florida, Gainesville, Florida

**Corresponding author:**
Juan P. Brito, MD, MS
Division of Endocrinology, Diabetes, Nutrition and Metabolism
Mayo Clinic
200 First St SW
Rochester, MN 55902
brito.juan@mayo.edu

Word count: 3361

# INTRODUCTION

Thyroid cancer incidence in the United States (U.S.) has tripled over the past three decades, rising from 5.5 per 100,000 people in 1990 to 13.4 per 100,000 in 2022.[1] It is now among the most common cancer in American adults aged 16-33, and the fifth most frequent cancer in women.[2] In 2025, an estimated 44,020 new cases are expected, largely driven by small papillary thyroid cancers (≤1.5 cm).[1,3] Despite the rise in incidence, mortality has remained low, leading to concerns about overdiagnosis due to detection of subclinical cancers unlikely to cause harm, which often result in unnecessary imaging, treatments, related adverse events, and significant financial burdens.[3-5]

A primary driver of this overdiagnosis is the escalating use of advanced medical imaging for indications unrelated to the thyroid. The number of CT scans performed annually in the U.S. has surged from 72 million in 2007 to nearly 93 million today, leading to a parallel increase in incidental thyroid findings (ITFs).[6,7] A recent meta-analysis confirms the scale of this issue, estimating the prevalence of ITF on CT scans to be 8.3%.[8] Each of these findings represents the potential start of a diagnostic cascade of further imaging, biopsies, and surgeries that may not offer clinical benefit.

Yet, despite their frequency and potential impact, incidental thyroid findings (ITFs) remain poorly characterized. Critical gaps persist in understanding their epidemiology, diagnostic evaluation, sonographic and radiographic features, and clinical outcomes. These uncertainties have contributed to inconsistent guidance: among the four major U.S. societies addressing thyroid nodules, two recommend ultrasound evaluation for all ITFs when a nodule is present, whereas others advocate selective evaluation based on nodule size, age, or comorbidities.[9-12] Progress has been limited because essential data is embedded within the unstructured narrative text of imaging reports within the electronic health record (EHR). Manual abstraction is neither feasible nor scalable, constraining prior research to small cohorts and limiting its generalizability.

Artificial intelligence (AI), especially natural language processing (NLP) techniques, offers a transformative solution to this long-standing data challenge.[13] By automating the analysis of unstructured text, NLP can unlock the rich clinical details embedded in radiology reports at an unprecedented scale.[14] However, a systematic review of prior work reveals that existing NLP algorithms for thyroid conditions have primarily focused on thyroid nodules using radiology reports. These algorithms have largely explored automation and classifications tasks (categorizing as benign vs malignant) rather than performing granular characterization of the nodular features or extracting radiology recommendations, both necessary to understand the drivers of the diagnostic cascade.[15]

Therefore, this study was designed to address these critical gaps. We developed and validated a novel, high-performance NLP pipeline to automate the identification of ITFs and, specifically, the detailed characterization of incidental thyroid nodules (ITNs). This characterization included extracting their radiologic features and management recommendations from a wide array of imaging modalities. Subsequently, we deployed this system across a large, multi-site healthcare network to quantify the association between these incidentally detected findings and the subsequent cascade of diagnostic interventions and thyroid cancer diagnoses.

## METHODS
**Data source and study population**

We conducted a retrospective study of adult patients who received care at Mayo Clinic sites in Rochester, Minnesota; Jacksonville, Florida; Scottsdale, Arizona, or the Mayo Clinic Health System (MCHS) in Minnesota and Wisconsin between July 1, 2017, and September 30, 2023. We used Current Procedural Terminology (CPT) and International Classification of Diseases (ICD) codes (Supplemental Table 1), to identify adult patients (≥18 years) who underwent imaging studies likely to incidentally capture the thyroid, including computed tomography (CT); magnetic resonance imaging (MRI) of the neck or chest; positron emission tomography combined with CT (PET/CT); carotid ultrasonography; parathyroid scintigraphy; octreotide scintigraphy; nuclear cardiac stress testing; and head or neck ultrasonography performed for non-thyroid indications. The study was approved by Mayo Clinic Institutional Review Board (23-003644).

To identify a cohort of newly detected ITFs, we applied strict exclusion criteria. We excluded patients with any prior diagnosis of thyroid nodule, thyroid cancer, hyperthyroidism, or a history of thyroidectomy or thyroid biopsy, using a comprehensive list of diagnostic and procedural codes (Supplemental Table 1). To ensure the absence of pre-existing thyroid disease, we required at least 12 months of continuous EHR data before the index imaging study. If patients had multiple qualifying imaging studies on the same day, we randomly selected one for inclusion to avoid duplication.

**Development of the NLP system**

To automate the identification and characterization of ITFs from unstructured imaging reports, we developed a two-stage, transformer-based NLP pipeline. We began with a convenience sample of 2,000 reports, manually reviewed to ensure representation across three categories: (1) no thyroid findings, (2) incidental non-nodular thyroid abnormalities (e.g., diffuse parenchymal changes), and (3) ITN. From this sample, we selected 300 reports for model development, maintaining the approximate distribution across categories (~50% no finding, ~45% ITN, ~5% non-nodular).

Each report was independently annotated by at least two clinically trained reviewers using MedTator, an annotation tool that supports data labeling for NLP use, to capture the presence and type of ITF (ITN vs non-nodular).[16] When applicable, reviewers also recorded nodule size and radiologic features consistent with those reported in routine imaging practice, including location (right, left, bilateral, isthmus), density, enhancement, calcifications, attenuation, and metabolic activity, as well as whether the radiologist recommended follow-up (e.g., ultrasound, additional imaging, or other testing). Annotation guidelines (Supplemental Table 2 & Supplemental Figure 1) were refined through iterative pilot testing until the inter-annotator agreement was high (κ > 0.8). To address false-positive classifications identified during preliminary model testing, an additional 60 reports were annotated and incorporated into the training set, improving model precision.

The NLP pipeline consisted of two sequential components. First, a classification module determined whether a report contained any ITF (Supplemental Figure 2). We fine-tuned multiple transformer models, including BioBERT, BioClinicalBERT, and GatorTron. Reports classified as positive were further categorized into ITNs or non-nodular findings using rule-based logic. Second, for reports containing ITNs, a named entity recognition (NER) module was developed to extract structured nodule attributes (Supplemental Figure 3). Three models, BERT-base, RoBERTa, and a domain-specific Medical-NER, were trained to identify ITN location, size, radiologic features, and follow-up recommendations (Supplemental Table 3). For the classification task (determining whether a report contained an ITF), BioClinicalBERT achieved the best performance (Testing F1-score 0.97). For the entity recognition task (extracting ITN attributes), Medical-NER performed best (Testing F1-score 0.81) (Supplemental Tables 4 and 5).

**Baseline variables and study outcomes**

For the overall cohort, we extracted baseline variables from the 12 months preceding the index imaging study. Demographic characteristics included age, sex, race, ethnicity, educational attainment, marital status, primary language, and insurance status. Employment and financial strain (as recorded within the social determinants of health) were also captured. Clinical variables included Body Mass Index (BMI) and comorbidity burden, measured using the Charlson Comorbidity Index.[17] Imaging-related variables included modality and body region examined. For patients with ITNs, additional radiologic features were extracted using the NLP pipeline in accordance with predefined categories: location (right, left, bilateral, isthmus), size, density, enhancement, calcifications, attenuation, metabolic activity, and metabolic distribution. We also captured whether the interpreting radiologist included a recommendation for further evaluation (e.g., ultrasound, additional imaging, or other testing).

The primary outcomes were the prevalence of ITFs and the proportion of patients with ITNs identified. Secondary outcomes included subsequent thyroid-directed evaluations and diagnoses following the index imaging study. Specifically, outcomes of interest were thyroid ultrasound, thyroid biopsy, thyroidectomy (partial or total), and new diagnoses of thyroid nodules or thyroid cancer. Thyroid ultrasound was considered a downstream outcome if performed within 180 days of the index imaging. Biopsy, thyroidectomy (partial or total), and new diagnoses of thyroid nodules or thyroid cancer were included if they occurred within 180 days after a thyroid ultrasound. Outcomes were ascertained using ICD and CPT codes (Supplemental Table 1).

For thyroid cancer ascertainment, a single diagnostic code has been shown to be insufficient for confirmation.[18] To reduce false positives, we applied stricter criteria. If thyroid surgery was performed, a single thyroid cancer diagnostic code after the surgery date was considered sufficient, given that surgery typically implies pathological confirmation. If thyroid surgery was not performed, at least three instances of a thyroid cancer diagnostic code were required to confirm the diagnosis. This approach increased specificity, consistent with the principle that multiple code occurrences enhance diagnostic certainty when relying on administrative data. Finally, a manual EHR review of confirmed cases was performed to extract tumor histology and size.

**Statistical analysis**

Descriptive statistics were used to summarize the variables of interest. Means and standard deviations (SD) were reported for continuous variables and counts and percentages were reported for categorical variables. Univariate logistic regression models were conducted to assess the associations of clinical characteristics and an incidental finding being present.

To identify factors associated with incidental findings, a multivariable logistic regression with Least Absolute Shrinkage and Selection Operator (LASSO) selection was performed, with a selection criterion of Schwarz Bayesian Criterion (SBC).[19] Candidate covariates included patient factors (sex, age, BMI, race, ethnicity, primary language, marital status, insurance, Charlson comorbidity index), clinician factors (specialty, sex, credentials) and imaging factors (modality, body location). Factors of interest selected were patient age, patient sex, BMI, modality description, body group and specialty. A significant interaction term modality description and body group was found to be highly significant (p<0.0001) assessed by the likelihood ratio chi-square test statistic, indicating that the effect of imaging modality on incidental findings varies by body region. Odds ratios and 95% confidence intervals (95% CI) were estimated for all predictors in the final model.

Model assumptions were verified. The Hosmer-Lemeshow goodness-of-fit test yielded a p-value of 0.15, indicating no evidence of poor model fit. The area under the receiver operating characteristic curve (AUC) was 0.74, suggesting good discrimination. Influential observations (Cook's distance > 0.5) were assessed and had minimal impact (< 10% coefficient change upon removal). Data management and statistical analysis were performed using SAS version 9.4 (SAS Institute, Inc., Cary, NC). Complete case analysis was used, as missing data for the variables of interest were minimal (< 7%).[20]

## RESULTS
### Study Cohort Demographics

The final cohort comprised 115,683 patients and corresponding unique radiology reports (Figure 1). The mean age was 56.8 years (SD 17.2), and the majority of patients were female (52.9%), White (91.0%), and not Hispanic (94.5%). Most patients had private insurance (51.5%) or Medicare (36.5%). The most common imaging modality was CT (62.5%), and the most frequent body location imaged was the chest (44.9%). Complete demographic details are provided in Table 1. Charlton comorbidity index and ordering clinician characteristics can be found in Supplemental Tables 6 & 7, respectively.

### Prevalence and Characteristics of Patients with Incidental Thyroid Findings

Among 115,683 patients, we identified 9,077 (7.8%) with an ITF (Table 1). In unadjusted analyses, ITFs were more frequent in women, older adults, and patients undergoing chest or neck imaging. Socioeconomic differences were also observed, with higher prevalence among patients with Medicare and lower prevalence among those with Medicaid or reported financial hardship. CT was the most common modality for detecting ITFs, although nuclear medicine and PET studies demonstrated higher crude odds of detection.

In multivariable analysis (Figure 2, Supplemental Table 8), several factors remained independently associated with ITFs. Women had nearly twice the odds compared with men (OR, 1.83; 95% CI, 1.75–1.92). Age was a strong predictor, with a 9.5% increase in odds per 5-year increment (OR, 1.09; 95% CI, 1.09–1.10). Higher BMI modestly increased risk (OR, 1.03 per 5-unit increment; 95% CI, 1.01–1.04).

Ordering a specialty was a key determinant. Compared with emergency medicine, orders from oncology (medical oncology: OR, 2.12; 95% CI, 1.90–2.37; hematology-oncology: OR, 2.05; 95% CI, 1.86–2.27 and internal medicine: OR, 1.64; 95% CI, 1.49–1.81, were most strongly associated with ITFs.

Modality and body region were highly associated with ITF detection and demonstrated a significant interaction (Supplemental Figure 4). Compared with a CT of the chest, a CT of the neck was 3.4 times more likely to detect an ITF (OR, 3.41; 95% CI, 3.20–3.63). This effect was even more pronounced with other modalities: a PET scan of the neck had 4.8 times the odds, and a nuclear medicine scan of the neck had over 25 times the odds of detecting an ITF (OR, 4.83; 95% CI, 1.32–17.89 and OR, 25.54; 95% CI, 9.77–66.00, respectively). Conversely, neck imaging via MRI (OR, 0.27; 95% CI, 0.10–0.75) and ultrasound (OR, 0.12; 95% CI, 0.02–0.83) was associated with lower odds compared to chest CT. Head imaging consistently showed lower odds of ITF detection across all modalities relative to chest CT. For chest imaging specifically, PET (OR, 5.90; 95% CI, 3.74–9.30) and nuclear medicine (OR, 3.07; 95% CI, 2.05–4.58) had higher odds of ITF detection than CT, whereas other modalities had lower odds.

**Radiologic Characteristics and Recommendations for Incidental Thyroid Nodules**

Among the 9,077 patients with an ITF, 8,426 (92.9%) had findings classified as ITNs (Table 2). Nodule location was the most frequently reported feature, documented in 6,504 cases (77.2%); the right lobe was most common (38.2%), followed by the left lobe (34.6%) and bilateral nodules (27.0%). Nodule size was reported in 3,689 cases (43.8%), with a mean diameter of 1.6 cm (SD, 1.2). Most nodules measured 1.1–2.0 cm (40.6%) or ≤1.0 cm (38.0%).

Other radiologic features were described far less often. Calcifications were reported in 14.3% of cases, while density (14.0%), attenuation (11.1%), and metabolic activity (8.9%) were infrequently documented. When present, nodules were most often described as low density (92.9%) or low attenuation (90.8%). Among reports noting metabolic activity, nearly half characterized it as ambiguous or unclear (46.9%). All reports mentioning calcifications simply recorded their presence (100%).

Radiologist recommendations were included in 26.7% of reports. When provided, follow-up ultrasound was most frequently advised (70.5%), followed by non-ultrasound imaging (10.0%) or other recommendations (19.5%).

**Clinical Outcomes Following Incidental Findings**

A follow up thyroid ultrasound was ordered in 20.2% (1,830) of patients with ITFs. A post-index thyroid nodule diagnosis was ascertained in 2,026 of 9,077 patients with an ITF (22.3%) compared with 676 of 106,606 patients without an ITF (0.6%). This led to a significant increase in downstream procedures. Patients with an ITF had substantially higher odds of undergoing thyroid biopsy (6.1% vs 0.1%; OR, 46.8; 95% CI, 39.0–56.2) and were more likely to receive a partial (OR, 87.6; 95% CI, 49.1–156.5) or total thyroidectomy (OR, 55.8; 95% CI, 31.3–99.3). Although a diagnosis of thyroid cancer remained uncommon overall, it occurred far more often among patients with an ITF: 109 cases (1.2%) compared with 21 cases (0.02%) in those without ITFs, corresponding to over 60-fold higher odds of cancer detection (OR, 61.7; 95% CI, 38.6–98.5) (Table 3).

Among patients with ITFs who were diagnosed with thyroid cancer, 96 of 109 cases (88.1%) were papillary thyroid carcinoma, 6 (5.5%) were oncocytic cell carcinoma, 4 (3.7%) were medullary, and 2 (1.8%) were follicular (Supplemental Table 9). By comparison, cancers detected in patients without ITFs were also predominantly papillary (19 of 21 cases, 90.5%), with only isolated cases of anaplastic (n=1, 4.8%) and oncocytic cell carcinoma (n=1, 4.8%). The mean tumor size was larger among cancers in the ITF group (19.7 mm; SD, 14.4) compared with those in the non-ITF group (12.8 mm; SD, 7.8; P = 0.0499).

**DISCUSSION**

In this large, multisite cohort of 115,683 adults undergoing imaging for non-thyroid indications, ITFs were common, present in 7.8%, nearly 1 in 13 patients. Most ITFs represented nodular disease (92.9%). Approximately one-fifth of patients with ITFs proceeded to a dedicated thyroid ultrasound, marking the first step of a pronounced diagnostic cascade: compared with patients without ITFs, those with ITFs had markedly higher odds of biopsy, thyroidectomy, and receipt of a thyroid cancer diagnosis. Most cancers detected were small papillary thyroid carcinomas, reinforcing concerns that incidental detection predominantly uncovers low-risk disease.

Our prevalence aligns closely with the 8.3% pooled estimate from a recent meta-analysis, providing external validation for our NLP-based methodology. However, the rate of follow-up thyroid ultrasound in our cohort (~20%) was lower than the 34.9% reported in pooled data.[8] This discrepancy likely reflects a combination of institutional practice norms, patient comorbidity burden, and variations in reporting culture. It illustrates that recognition of an incidentaloma does not uniformly translate into cascade initiation but is instead filtered through local practice patterns and patient context.

When cascades were triggered, however, the effect was substantial: patients with ITFs had >40-fold higher odds of biopsy and >60-fold higher odds of a thyroid cancer diagnosis compared with those without ITFs. Prior studies similarly show that once identified, ITNs often prompt biopsy in up to one-third of cases and surgery in 5%–10%, despite malignancy rates of only 2%–5% overall. In our study, the malignancy rate was 1.2%, lower than prior estimates, yet the histologic profile was consistent—dominated by papillary carcinomas ≤2 cm.[21,22] Taken together, these findings demonstrate that while entry into the cascade is variable, once initiated it reliably channels patients toward the detection of small, low-risk tumors, underscoring the central role of ITFs as a driver of thyroid cancer overdiagnosis in the United States.

Multivariable modeling clarified the demographic and imaging determinants of ITFs detection. Female sex and older age were strong predictors, consistent with the epidemiology of thyroid nodules.[23,24] Detection varied by modality and body region, with PET and nuclear studies showing the highest odds and MRI the lowest, while neck imaging was far more likely than chest to yield an ITFs. These patterns mirror the heterogeneity observed in prior single-center cohorts, where chest CT prevalence typically ranged from 5–7% and neck CT estimates exceeded 15%.[8] Our results extend these observations by demonstrating a significant modality-by-region interaction effect across a large, multi-institutional population.

Real-world reporting of radiologic features was inconsistent. Only 44% of ITNs had a documented size, and fewer than 15% had descriptors such as calcifications, density, or metabolic activity. This gap parallels concerns raised in narrative reviews and the ACR white paper, which emphasize that CT and MRI lack the fine-grained features of ultrasound.[10,25,26] Our results show how reliance on incomplete reporting forces clinicians to make cascade decisions based largely on size, a blunt tool that increases risk of both over- and under-evaluation.[27]

This study should be interpreted considering its limitations. First, although the cohort is large and drawn from multiple sites, all data comes from a single healthcare system, and the patient population and institutional practices may not be fully generalizable to other settings. Second, the variable quality of documentation in radiology reports limited our ability to analyze which features most strongly predict malignancy or downstream interventions, both a study limitation and a reflection of current reporting practice. Third, although the NLP models performed well, misclassification is possible and could modestly influence prevalence and outcome estimates. Finally, this analysis does not provide a definitive characterization of incidental thyroid nodules; such work would require direct review of images rather than radiology reports. By relying on radiologist documentation, our findings may underestimate the true frequency of nodules and be biased toward those considered significant enough to report.

This study advances understanding of incidental thyroid findings by linking prevalence with downstream outcomes, creating opportunities to study the drivers of cascade initiation, track trends over time, and evaluate interventions across health systems and countries, provided the NLP pipeline is validated in diverse reporting environments. Variation in follow-up ultrasound highlights the need to determine whether limited use reflects appropriate restraint or missed opportunities for care. For practice, including guideline development, these findings highlight the importance of standardized reporting; the creation of a structured lexicon for CT and other cross-sectional imaging—analogous to the American College of Radiology Thyroid Imaging Reporting and Data System (ACR TI-RADS) for ultrasound—could reduce variability, facilitate international comparisons, and promote cascade-aware management.[28]

By developing and deploying a high-performance NLP system, this study unlocked clinical insights from unstructured radiology reports at scale. Incidental thyroid findings were common and, when acted upon, strongly associated with a diagnostic cascade that predominantly uncovered small, low-risk papillary carcinomas, providing large-scale evidence that ITFs are a primary driver of thyroid cancer overdiagnosis. These findings highlight the urgent need for standardized reporting of cross-sectional imaging and support efforts to reduce unnecessary diagnostic cascades while preserving detection of clinically meaningful disease.

**Author Contributions:** Drs Larios and Brito had full access to all the data in the study and take responsibility for the integrity of the data and the accuracy of the data analysis. Drs Larios and Borras-Osorio contributed equally as co–first authors.

**Concept and design:** Larios, Borras-Osorio, Toro-Tobon, Loor-Torres, Wu, Singh Ospina, Brito.
**Acquisition, analysis, or interpretation of data:** Larios, Borras-Osorio, Wu, Toro-Tobon, Cabezas, Claros, Guevara Maldonado, Vilatuna Andrango, Lizarazo, Mateo Chavez, Proano, Soto Jacome, Fan, Loor-Torres, Al Zahidy, Ponce-Ponte, Branda.
**Drafting of the manuscript:** Larios, Borras-Osorio, Wu, Claros, Toro-Tobon, Guevara Maldonado, Vilatuna Andrango, Lizarazo, Alzamora, Montero, Brito.
**Critical review of the manuscript for important intellectual content:** Larios, Borras-Osorio, Toro-Tobon, Proano, Soto Jacome, Alzamora, Montero, Ponce-Ponte, Mateo Chavez, Fan, Loor-Torres, Al Zahidy, Claros, Cabezas, Wu, Branda, Singh Ospina.
**Statistical analysis:** Wu, Branda.
**Administrative, technical, or material support:** Ponce-Ponte, Fan, Loor-Torres, Al Zahidy, Cabezas, Alzamora, Montero, Wu.
**Supervision:** Larios, Borras-Osorio, Wu, Ponce-Ponte, Singh Ospina, Brito.

**Conflict of Interest Disclosures:** Dr Toro-Tobon reported personal fees from Immunovant outside the submitted work. Dr Fan reported grants from the National Institutes of Health during the conduct of the study. Dr Branda reported grants from the National Cancer Institute during the conduct of the study. No other disclosures were reported. Funding/Support: Dr Singh Ospina was supported by the National Cancer Institute of the National Institutes of Health under award K08CA248972. Dr Brito was supported by the National Cancer Institute of the National Institutes of Health under award R37CA272473.

**Data Sharing Statement:** Deidentified participant data that support the findings of this study are available from the corresponding author at brito.juan@mayo.edu upon reasonable request.

**Table 1.** Demographics - Incidental Findings vs. Non-Incidental

|  | Incidental Findings | | | |
|---|---|---|---|---|
|  | No<br>(N=106606) | Yes<br>(N=9077) | Total<br>(N=115683) | Odds Ratio (95% CI)[1] |
| **Gender, n (%)** | | | | |
| Female | 55353 (51.9%) | 5860 (64.6%) | 61213 (52.9%) | 1.69 (1.62, 1.77) |
| Male | 51219 (48.1%) | 3208 (35.4%) | 54427 (47.1%) | Ref |
| **Age (years)** | | | | |
| Mean (SD) | 56.4 (17.25) | 61.2 (15.49) | 56.8 (17.2) | 1.09 (1.08, 1.10) |
| **Age, n (%)** | | | | |
| 18-54 | 43524 (40.8%) | 2670 (29.4%) | 46194 (39.9%) | Ref |
| 55-65 | 23354 (21.9%) | 2089 (23.0%) | 25443 (22.0%) | 1.46 (1.37, 1.55) |
| 65+ | 39728 (37.3%) | 4318 (47.6%) | 44046 (38.1%) | 1.77 (1.69, 1.86) |
| **Race, n (%)** | | | | |
| Black | 4533 (4.3%) | 492 (5.5%) | 5025 (4.4%) | 1.30 (1.18, 1.43) |
| Asian | 2588 (2.5%) | 265 (3.0%) | 2853 (2.5%) | 1.22 (1.08, 1.39) |
| White | 95553 (91.1%) | 8006 (89.6%) | 103559 (91.0%) | Ref |
| Other | 2209 (2.1%) | 168 (1.9%) | 2377 (2.1%) | 0.91 (0.78, 1.06) |
| **Ethnicity, n (%)** | | | | |
| Hispanic | 5847 (5.6%) | 370 (4.2%) | 6217 (5.5%) | 0.73 (0.66, 0.82) |
| Not Hispanic | 98464 (94.4%) | 8496 (95.8%) | 106960 (94.5%) | Ref |
| **BMI** | | | | |
| Mean (SD) | 29.1 (7.05) | 29.4 (7.02) | 29.1 (7.1) | 1.03 (1.01, 1.04) |
| **Primary Language, n (%)** | | | | |
| Non-English | 2728 (2.6%) | 293 (3.2%) | 3021 (2.6%) | 1.27 (1.12, 1.44) |
| English | 103786 (97.4%) | 8775 (96.8%) | 112561 (97.4%) | Ref |
| **Marital status, n (%)** | | | | |
| Married/Life Partner | 68019 (64.1%) | 5914 (65.6%) | 73933 (64.2%) | Ref |
| Divorced/Separated | 9162 (8.6%) | 765 (8.5%) | 9927 (8.6%) | 0.96 (0.89, 1.04) |
| Widowed | 7000 (6.6%) | 892 (9.9%) | 7892 (6.9%) | 1.47 (1.36, 1.58) |
| Single | 21922 (20.7%) | 1448 (16.1%) | 23370 (20.3%) | 0.76 (0.72, 0.81) |
| **Education, n (%)** | | | | |
| No high school degree | 2349 (3.0%) | 178 (2.6%) | 2527 (2.9%) | 0.94 (0.80, 1.10) |
| Highschool Degree/GED | 29819 (37.5%) | 2412 (35.7%) | 32231 (37.3%) | Ref |
| Some college/associate degree | 13449 (16.9%) | 1140 (16.9%) | 14589 (16.9%) | 1.05 (0.97, 1.13) |
| Bachelor's degree | 18742 (23.6%) | 1632 (24.1%) | 20374 (23.6%) | 1.08 (1.01, 1.15) |
| Post-Graduate Degree | 15182 (19.1%) | 1397 (20.7%) | 16579 (19.2%) | 1.14 (1.06, 1.22) |
| **Employment, n (%)** | | | | |
| Disabled | 5208 (6.8%) | 312 (4.8%) | 5520 (6.6%) | 0.82 (0.72, 0.92) |
| Employed | 34097 (44.4%) | 2505 (38.5%) | 36602 (44.0%) | Ref |
| Retired | 31063 (40.5%) | 3204 (49.2%) | 34267 (41.2%) | 1.40 (1.33, 1.48) |
| Unemployed | 6373 (8.3%) | 488 (7.5%) | 6861 (8.2%) | 1.04 (0.94, 1.15) |
| **Payor, n (%)** | | | | |
| Private | 55416 (52.0%) | 4196 (46.2%) | 59612 (51.5%) | Ref |
| Medicare | 38225 (35.9%) | 4044 (44.6%) | 42269 (36.5%) | 1.40 (1.34, 1.46) |
| Medicaid | 8817 (8.3%) | 514 (5.7%) | 9331 (8.1%) | 0.77 (0.70, 0.85) |
| Other Govt Programs | 2272 (2.1%) | 157 (1.7%) | 2429 (2.1%) | 0.91 (0.77, 1.08) |
| Self-Pay | 1876 (1.8%) | 166 (1.8%) | 2042 (1.8% | 1.17 (0.99, 1.37) |
| **Financial risk, n (%)** | | | | |
| Not hard | 58317 (83.9%) | 5205 (85.7%) | 63522 (84.1%) | Ref |
| Somewhat hard | 7739 (11.1%) | 615 (10.1%) | 8354 (11.1%) | 0.89 (0.82, 0.97) |
| Hard/Very Hard | 3442 (5.0%) | 257 (4.2%) | 3699 (4.9%) | 0.84 (0.74, 0.95) |
| **Imaging Modality, n (%)** | | | | |
| COMPUTED TOMOGRAPHY | 65130 (61.1%) | 7212 (79.5%) | 73242 (62.5%) | Ref |
| MAGNETIC RESONANCE | 24840 (23.3%) | 598 (6.6%) | 25438 (22.0%) | 0.22 (0.20, 0.24) |
| NUCLEAR MEDICINE | 2248 (2.1%) | 494 (5.4%) | 2742 (7.2%) | 1.99 (1.80, 2.19) |
| POSITRON EMISSION TOMOGRAPHY | 7640 (7.2%) | 674 (7.4%) | 8314 (7.2%) | 0.80 (0.73, 0.87) |
| CAROTID ULTRASOUND | 6748 (6.3%) | 99 (1.1%) | 6847 (5.9%) | 0.13 (0.11, 0.16) |
| **Body Location, n (%)** | | | | |
| Head | 16794 (15.8%) | 389 (4.3%) | 17183 (14.9%) | 0.24 (0.22, (0.27) |
| Neck | 15559 (14.6%) | 2367 (26.1%) | 17926 (15.5%) | 1.60 (1.52, 1.69) |
| Chest | 47468 (44.5%) | 4505 (49.6%) | 51973 (44.9%) | Ref |
| Mixed | 26785 (25.1%) | 1816 (20.0%) | 28601 (24.7%) | 0.71 (0.68, 0.76) |

[1]Complete case analysis with univariate logistic regression where dependent variable is incidental finding (Yes vs. No) and independent variable is the patient demographics; [2]Age increments of 5 years; [3]BMI increments of 5 units;

**Table 2.** Imaging Features of Incidentally Detected Thyroid Nodules on imaging modalities Scans (N=8426)

| Imaging Feature Category | Subcategory | Nodules, N (%) |
|---|---|---|
| **Location** | | 6504 (77.2%) |
| | Bilateral | 1755 (27%) |
| | Isthmus | 9 (0.1%) |
| | Left | 2253 (34.6%) |
| | Right | 2487 (38.2%) |
| **Recommendation** | | 2249 (26.7%) |
| | Recommendation for ultrasound | 1585 (70.5%) |
| | Recommendation for non-ultrasound imaging | 226 (10%) |
| | Other recommendation | 438 (19.5%) |
| **Size** | | 3689 (43.8%) |
| | Mean (SD) | 1.6 (1.2) |
| | ≤ 1 cm | 1402 (38.0%) |
| | 1.1 - 2 cm | 1497 (40.6%) |
| | 2.1 – 3 cm | 468 (12.7%) |
| | 3.1 – 4 cm | 202 (5.5%) |
| | > 4 cm | 120 (3.3%) |
| **Density** | | 1178 (14.0%) |
| | Low density | 1095 (92.9%) |
| | Heterogeneous density | 36 (3.1%) |
| | High density | 28 (2.4%) |
| | Missing | 19 |
| **Enhancement** | | 322 (3.8%) |
| | Low enhancement | 1 (0.3%) |
| | High enhancement | 2 (0.6%) |
| | Heterogeneous enhancement | 16 (5.0%) |
| | Enhancing | 303 (94.1%) |
| **Calcifications** | | 1208 (14.3%) |
| | Calcified | 1208 (100%) |
| **Attenuation** | | 935 (11.1%) |
| | Low attenuation | 849 (90.8%) |
| | Heterogeneous attenuation | 43 (5.1%) |
| | High attenuation | 12 (1.4%) |
| | Attenuation | 31 (3.7%) |
| **Metabolic Activity** | | 753 (8.9%) |
| | Ambiguous/Unclear | 353 (46.9%) |
| | Hypermetabolic | 295 (39.2%) |
| | Low Metabolic | 55 (7.3%) |
| | Mid Metabolic | 10 (1.3%) |
| | Non-Metabolic | 31 (4.1%) |
| | Physiological | 9 (1.2%) |
| **Metabolic Distribution** | | 759 (9.0%) |
| | Diffuse | 145 (19.1%) |
| | Focal/Localized | 107 (14.1%) |
| | Not described | 507 (66.8%) |
| **Other** | | 781 (9.3%) |

**Table 3.** Clinical outcomes: Incidental Findings vs. Non-Incidental

|  | Incidental Findings | | |
|---|---|---|---|
|  | No (N=106606) | Yes (N=9077) | Odds Ratio (95% CI)[1] |
| **Thyroid Nodule, n (%)** | | | |
| No | 105930 (99.4%) | 7051 (77.7%) | |
| Yes | 676 (0.6%) | 2026 (22.3%) | 45.0 (41.1, 49.3) |
| **Thyroid biopsy, n (%)** | | | |
| No | 106458 (99.9%) | 8522 (93.9%) | |
| Yes | 148 (0.1%) | 555 (6.1%) | 46.8 (39.0, 56.2) |
| **Partial Thyroidectomy, n (%)** | | | |
| No | 106593 (100%) | 8981 (98.9%) | |
| Yes | 13 (0.0%) | 96 (1.1%) | 87.6 (49.1, 156.5) |
| **Total Thyroidectomy, n (%)** | | | |
| No | 106592 (100%) | 9011 (99.3%) | |
| Yes | 14 (0.0%) | 66 (0.7%) | 55.8 (31.3, 99.3) |
| **Thyroid Cancer, n (%)** | | | |
| No | 106585 (100%) | 8968 (98.8%) | |
| Yes | 21 (0.0%) | 109 (1.2%) | 61.7 (38.6, 98.5) |

[1]Univariate Logistic Regression where clinical outcomes are the dependent variable and incidental finding (Yes vs. No) is the independent variable.

**Figure 1.** Cohort identification. *Training and testing sets were constructed using random sampling, combined with targeted data enrichment to ensure sufficient representation of thyroid incidental findings.

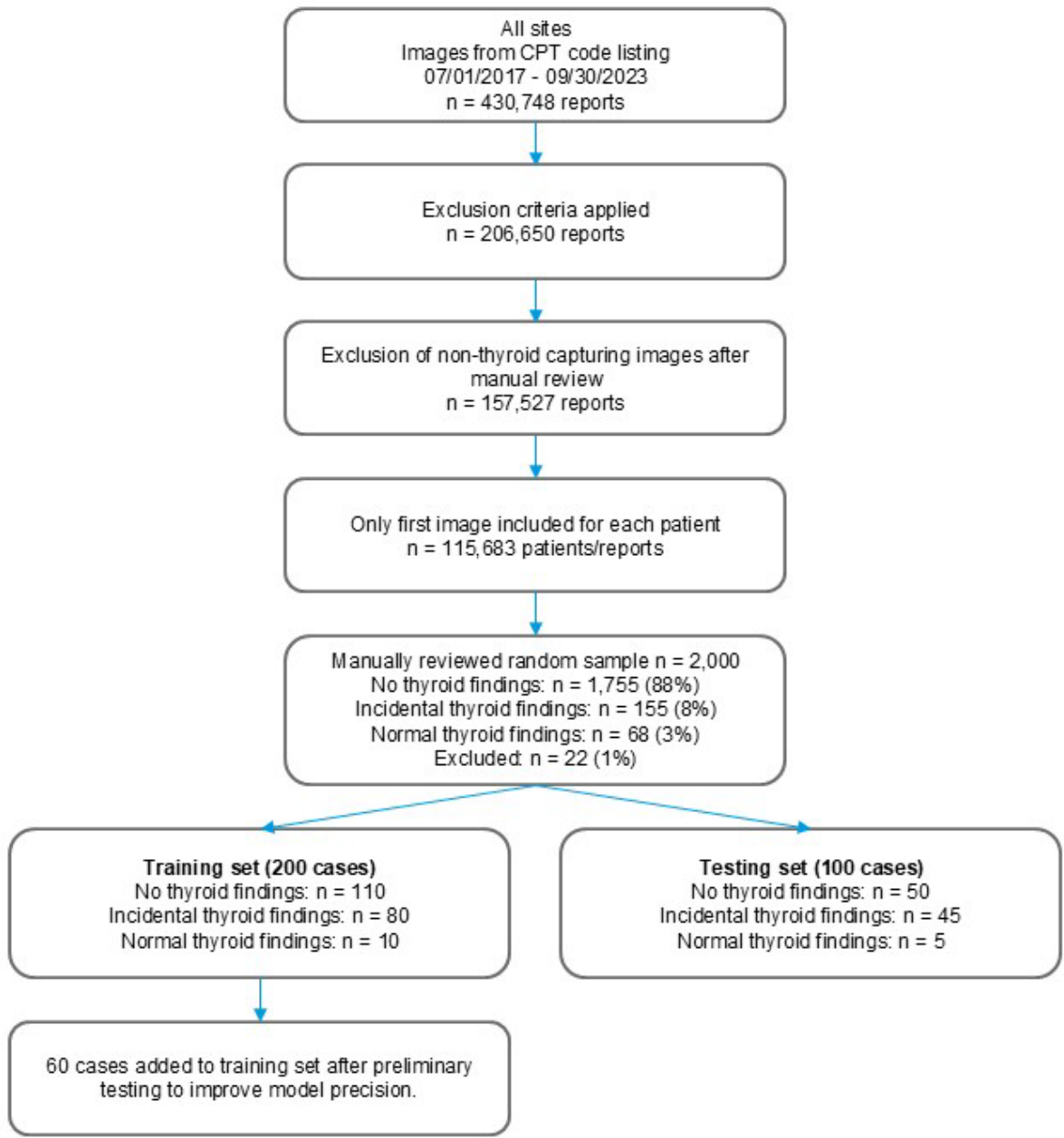

**Figure 2.** Multivariable logistic regression model with dependent variable of INF vs not adjusted by Sex, patient age in years in 5-year increments, patients BMI in 5-unit increments, ordering clinician specialty, Modality, body imaging location and the interaction of these two variables. The forest plot uses a logarithmic scale for the x-axis to effectively display a wide range of odds ratios (0.5 to 66) while maintaining clarity and proportionality for both small and large values.

BMI: Body Mass Index; ED: Emergency Department; FM: Family Medicine; IM: Internal Medicine; MRI: Magnetic Resonance Imaging; CT: Computed Tomography; PET: Positron Emission Tomography.

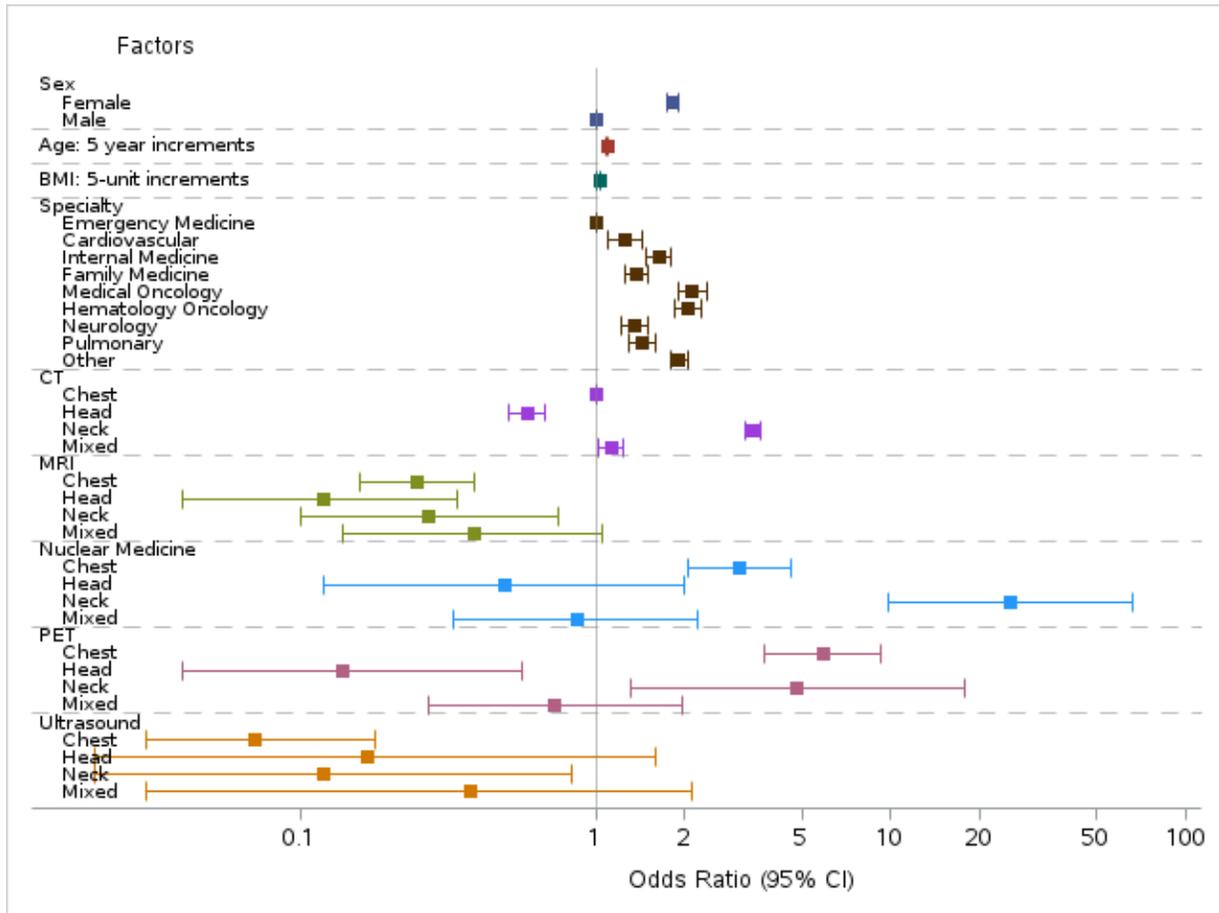

## SUPPLEMENTAL MATERIALS

**Supplemental Table 1.** Inclusion criteria, exclusion criteria, and list of codes.

| Inclusion and Exclusion criteria |
| --- |
| Inclusion: |
| • Adult patients who underwent one or more imaging modalities (see below) in Mayo Clinic Rochester, Jacksonville, Scottsdale or Midwest Healthcare System from 07/01/2017 to 09/30/2023. |
| • Only the first image is included. |
| Exclusion: |
| • Previous diagnosis of thyroid cancer, thyroid nodule, partial or total thyroidectomy, thyroid biopsy or hyperthyroidism (see below). |
| • No research authorization. |
| • Less than 12 months of previous visits/care at Mayo Clinic. |

| Image Description | Code (CPT) |
| --- | --- |
| PET and PET/CT | G0210-G0234, G0252-G0254, G0330-G0336, 78608, 78810, 78811-78813, 78814-78816 |
| CT neck | 70480-70492 |
| CT chest | 71250, 71260, 71270 |
| CT cervical spine | 72125-71127 |
| CT angiography neck | 70498 |
| CT angiography chest | 71275 |
| MRA neck or chest | 70547, 70549, 70540, 70543 |
| MRI cervical spine | 72141, 72142, 72156 |
| MRI Orbit, face, neck | 70540, 70542, 70543 |
| MRI chest | 71550, 71551, 71552 |
| Ultrasound Carotid | 93880 |
| Face, head and neck soft tissue | 76536 |
| Parathyroid scan | 78070 |
| Octreotide scan | 78804, 78999 |
| Nuclear cardiac stress test | 78452, 93017 |
| **Diagnosis/Procedure Description** | **Code** |
| Thyroid Cancer | ICD-9: 193 |
|  | ICD-10: C73 |
| Cystic Thyroid Nodule | ICD-9: 246.2 |
| Thyroid Nodule | ICD-9: 226, 241.0, 241.1, 240.0, 241.9, 246.2 |
|  | ICD-10: D34 E01.1 D44.0 E04.1 E04.2 |
|  | HCPCS: G9552 |
| Thyroidectomy | CPT: 60212, 60225, 60240, 60252, 60254, 60270, 60271 |
| Partial Thyroidectomy | CPT: 60210, 60220 |
| Thyroid nodule biopsy | CPT: 60001, 60100, 60300 |
| Hyperthyroidism | ICD-9: 242.01, 242.10, 242.11, 242.20, 242.21, 242.30, 242.31, 242.40, 242.41, 242.80, 242.81, 242.90, 242.91, 245.0, 245.0, 245.1, 245.3, 245.4 |
|  | ICD-10: E05.00, E05.10, E05.20 |

**Supplemental Table 2.** Annotation Dictionary

| Category | Description | Example |
|---|---|---|
| **Incidental finding** | Header of the reports<br>**Annotate to be able to save all the reports (with and without incidental findings) | |
| **Normal findings** | Words that describe observation of normal thyroid gland | Without category options, no relevant for analysis.<br>Examples: normal, unremarkable |
| **Thyroid** | **Presence of the word thyroid in the radiology report, to start the screening workflow.**<br>** Words "goiter" and "thyromegaly" are going to be annotated under this category just the first time they are mentioned in the report, and just in case the word "thyroid" is not written in the report. If they appear more times, the following will be annotated as "type of finding" | ✓ **Thyroid (e.g., thyroid, thyroid gland)**<br>✓ **Goiter**<br>✓ **Thyromegaly** |
| **Type of finding** | The classification of the incidental finding in the thyroid gland. | ✓ Nodular (e.g., nodule, nodularity, nodular, multinodular, multinodular goiter, multinodular enlargement, goitrous enlargement, lesion, lesions, cyst, cysts, mass)<br>✓ Non-nodular (e.g., atrophy, atrophic, foci, focus, congenitally absent, remnant, enlargement, enlarged, calcification, calcified, ectopic tissue, thyroglossal duct, thyromegaly) |
| **Number of findings** | Number of thyroid incidental findings described either in a qualitative or quantitative form. | ✓ Single (e.g., single, one, 1, sole, unique)<br>✓ Multiple |
| **Size** | Size of the thyroid finding, reported either as a quantitative or qualitative characteristic (e.g., 4.6 cm, 12 mm, tiny, small, massive) | |
| **Location** | Refers to the specific location within the thyroid gland where the incidental finding is located. | ✓ Right lobe (e.g., right, right lobe, right thyroid lobe, right thyroid)<br>✓ Left lobe (e.g., left, left lobe, left thyroid lobe, left thyroid)<br>✓ Isthmus<br>✓ Bilateral |
| **Radiological characteristics** | Other radiological characteristics of the thyroid incidental finding, specific of the imaging modality used. | ✓ Density (e.g., hypodense, hyperdense, low dense, low density)<br>✓ Attenuation (e.g., hypoattenuating, hyperattenuating, heterogeneous attenuation, low attenuating)<br>✓ Metabolic activity (e.g., hypermetabolic, hypermetabolic, FDG activity, FDG uptake, SUV)<br>✓ Enhancement (e.g., hypoenhancing, hyperenhancing, normal enhancement)<br>✓ Calcification, calcifications, calcified |
| **Associated findings** | Other findings in the imaging report outside the thyroid but related to the thyroid finding. | ✓ Trachea invasion (e.g., tracheal FDG uptake, tracheal mass/lesion, trachea erosion, trachea invasion, tracheal wall thickening)<br>✓ Trachea non-invasion (e.g., narrowing of the upper trachea, tracheal deviation)<br>✓ Carotid/Jugular invasion (e.g., invasion of the vessel walls, irregular margin of the wall, loss of vascular wall layers, contour irregularities)<br>✓ Carotid/Jugular non-invasion (e.g., carotid/ jugular stenosis, lateral deviation)<br>✓ Esophagus invasion (e.g., esophagus FDG uptake, esophagus mass/lesion, esophagus erosion, esophagus invasion, esophagus wall thickening)<br>✓ Esophagus non-invasion (e.g., esophageal narrowing, esophagus deviation)<br>✓ Nerve invasion (e.g., )<br>✓ Nerve non-invasion (e.g., dysfunction of the recurrent pharyngeal nerve, displaced nerve)<br>✓ Vocal cords invasion (e.g., mass/irregularity of vocal cord, distortion of laryngeal soft tissue, increased radiotracer uptake)<br>✓ Vocal cords non-invasion (e.g., vocal cord not medialized, vocal cord paralysis) |

| | | | |
|---|---|---|---|
| | | ✓ | Cartilage invasion (e.g., erosion/destruction of thyroid/cricoid cartilage) |
| | | ✓ | Cartilage non-invasion (e.g., ) |
| | | ✓ | Bone invasion (e.g., osteolytic/osteoblastic bone lesion, pathological fracture, focal increased uptake/metabolic activity) |
| | | ✓ | Bone non-invasion (e.g., ) |
| | | ✓ | Muscle invasion (e.g., muscle mass/lesion, muscle thickening, focal increased uptake/metabolic activity) |
| | | ✓ | Muscle non-invasion (e.g., ) |
| | | ✓ | Other (e.g., thyroid extend into the thoracic inlet) |
| **Radiology classification** | Classification of the thyroid incidental finding, according to its characteristics in the image, given by the radiologist in the report. | ✓ | Benign (e.g., benign, likely benign, insignificant by size criteria, insignificant) |
| | | ✓ | Malignant (e.g., malignant, pathological, likely malignant, likely pathological, likely cancer) |
| | | ✓ | Indeterminate |
| **Recommendation** | Recommendations given by the radiologist related to the thyroid incidental finding (e.g., [image/biopsy] recommended, follow-up recommended, correlation recommended, nonemergent ultrasound recommended, consider [image/biopsy]) | \multicolumn{2}{l|}{Recommendation for thyroid incidental finding (yes/no)<br>- Clinical input needed<br>- Further imaging needed} |

**Supplemental Table 3.** Additional Information about applied models.

| Model | Description | Pre-Training Corpus | Key Features |
|---|---|---|---|
| **BERT-base** | Foundational version of BERT, designed to capture deep contextual relationships within text through bidirectional training. Used as a baseline model in this study. | BooksCorpus and English Wikipedia | General-purpose training; lacks domain-specific pre-training for medical text. |
| **RoBERTa** | An optimized version of BERT with improved training parameters, such as removal of the next-sentence prediction objective and use of larger mini-batches and learning rates. Well-suited for nuanced language comprehension tasks. | Extended corpus including books and English Wikipedia | Enhanced understanding of language dynamics; lacks biomedical-specific pre-training. |
| **Medical-NER** | A specialized model fine-tuned from DeBERTa, which incorporates disentangled attention mechanisms to better understand word relationships. Specifically trained for biomedical tasks using domain-specific corpora. | PubMed dataset | Domain-specific training for medical terminology and text structure; excels in identifying clinical entities within thyroid radiology reports. |

**Supplemental Table 4.** Comparative performance of the models in the identification of ITNs.

| Model | Accuracy | Precision | Recall | F1-score |
|---|---|---|---|---|
| **BERT-base** | 0.66 | 0.33 | 0.50 | 0.40 |
| **BioBERT** | 0.94 | 0.92 | 0.95 | 0.93 |
| **Bio_ClinicalBERT** | **0.97** | **0.96** | **0.98** | **0.97** |
| **Gatortron-base** | 0.81 | 0.82 | 0.86 | 0.81 |

**Supplemental Table 5.** Comparative performance of the models in the characterization of ITNs.

| Model | Entity | Precision | Recall | F1 score | Accuracy |
|---|---|---|---|---|---|
| **BERT-base** | Location | 0.64 | 0.78 | 0.70 | 0.82 |
| | Radiological Characteristics | 0.83 | 0.83 | 0.83 | |
| | Size | 0.75 | 0.90 | 0.82 | |
| | Recommendation | 0.50 | 0.50 | 0.50 | |
| | Average | 0.75 | 0.82 | 0.78 | |
| **Medical-NER** | Location | 0.89 | 0.89 | 0.89 | **0.85** |
| | Radiological Characteristics | 0.82 | 0.75 | 0.78 | |
| | Size | 0.64 | 0.90 | 0.75 | |
| | Recommendation | 1.00 | 1.00 | 1.00 | |
| | **Average** | **0.78** | **0.85** | **0.81** | |
| **RoBERTa-base** | Location | 0.70 | 0.64 | 0.67 | 0.60 |
| | Radiological Characteristics | 0.38 | 0.56 | 0.45 | |
| | Size | 0.35 | 0.70 | 0.47 | |
| | Recommendation | 0.50 | 0.33 | 0.40 | |
| | Average | 0.47 | 0.60 | 0.52 | |

**Supplemental Table 6.** Charlson comorbidity index - Incidental Findings vs Non-Incidental.

|  | Incidental Findings | | |
|---|---|---|---|
|  | No (N=106606) | Yes (N=9077) | Odds Ratio (95% CI)[1] |
| **Charlson Comorbidity Count, n (%)** | | | |
| 0 | 75430 (70.8%) | 6212 (68.4%) | Ref |
| 1 | 20820 (19.5%) | 1918 (21.1%) | 1.12 (1.06, 1.18) |
| 2 | 6425 (6.0%) | 578 (6.4%) | 1.09 (1.00, 1.19) |
| 3+ | 3931 (3.7%) | 369 (4.1%) | 1.14 (1.02, 1.27) |
| **Sum of Diseases: Mean (SD)** | 0.4 (0.86) | 0.5 (0.87) | 1.04 (1.02, 1.07) |
| **Severity Weighted Sum: Mean (SD)** | 0.7 (1.47) | 0.8 (1.59) | 1.04 (1.03, 1.06) |
| **Age weighted severity sum: Mean (SD)** | 2.2 (2.16) | 2.6 (2.22) | 1.09 (1.08, 1.10) |
| **Comorbidities Present:** | | | |
| Aids, n (%) | 90 (0.1%) | 4 (0.0%) | |
| Cerebrovascular Disease, n (%) | 1317 (1.2%) | 88 (1.0%) | |
| CHF, n (%) | 2764 (2.6%) | 220 (2.4%) | |
| CPD, n (%) | 6454 (6.1%) | 484 (5.3%) | |
| Dementia, n (%) | 818 (0.8%) | 62 (0.7%) | |
| Diabetes, n (%) | 7288 (6.8%) | 662 (7.3%) | |
| Diabetes Organ Damage, n (%) | 2978 (2.8%) | 264 (2.9%) | |
| Hemiplegia, n (%) | 184 (0.2%) | 7 (0.1%) | |
| Liver Disease, n (%) | 710 (0.7%) | 38 (0.4%) | |
| Metastatic Solid Tumor, n (%) | 1273 (1.2%) | 162 (1.8%) | |
| Mild Liver Disease, n (%) | 2643 (2.5%) | 191 (2.1%) | |
| Other Cancer, n (%) | 9936 (9.3%) | 1156 (12.7%) | |
| Peripheral Vascular Disease, n (%) | 2978 (2.8%) | 252 (2.8%) | |
| Renal Disease, n (%) | 5409 (5.1%) | 543 (6.0%) | |
| Rheumatologic Disease, n (%) | 1793 (1.7%) | 146 (1.6%) | |
| Ulcer, n (%) | 313 (0.3%) | 28 (0.3%) | |

[1]Univariate logistic regression.

**Supplemental Table 7.** Ordering clinician characteristics.

|  | Total (N=8897) | Clinicians with at least 1 Incidental finding (N=3365) |
|---|---|---|
| **Specialty, n (%)** | | |
| Emergency Medicine | 613 (7.1%) | 417 (12.6%) |
| Pulmonary Medicine | 226 (2.6%) | 157 (4.7%) |
| Hematology Oncology | 468 (5.4%) | 240 (7.2%) |
| Medical Oncology | 294 (3.4%) | 160 (4.8%) |
| Neurology | 676 (7.8%) | 259 (7.8%) |
| Cardiovascular | 666 (7.7%) | 186 (5.6%) |
| Internal Medicine | 841 (9.7%) | 310 (9.3%) |
| Family Medicine | 1074 (12.4%) | 390 (11.7%) |
| Other | 3805 (43.9%) | 1201 (36.2%) |
| **Gender, n (%)** | | |
| Female | 4154 (47.9%) | 1516 (45.7%) |
| Male | 4526 (52.1%) | 1804 (54.3%) |
| **Credentials, n (%)** | | |
| Physicians | 6383 (73.3%) | 2528 (76.0%) |
| Mid-Level Practitioners | 1802 (20.7%) | 659 (19.8%) |
| Nurses | 100 (1.1%) | 28 (0.8%) |
| Advanced practice providers | 234 (2.7%) | 58 (1.7%) |
| Associate clinicians | 41 (0.5%) | 10 (0.3%) |
| Others | 151 (1.7%) | 42 (1.3%) |
| **Patients[1]: Mean (Std)** | 13.0 (24.4) | 2.7 (3.2) |

[1]Patient count who meet criteria and are part of cohort. Mean number of patients per clinician.

**Supplemental Table 8.** Odds ratios for multivariable analysis.

| Characteristic | Levels | Odds Ratio (95% CI) |
|---|---|---|
| **Sex** | Female | 1.83 (1.75, 1.92) |
| | Male | Ref |
| **Patient Age** | 5-year increments | 1.095 (1.09, 1.10) |
| **BMI** | 5-unit increments | 1.03 (1.01, 1.04) |
| **Specialty** | Emergency Medicine | Ref |
| | Cardiovascular | 1.25 (1.10, 1.43) |
| | Internal Medicine | 1.64 (1.49, 1.81) |
| | Family Medicine | 1.38 (1.25, 1.51) |
| | Medical Oncology | 2.12 (1.90, 2.37) |
| | Hematology Oncology | 2.05 (1.86, 2.27) |
| | Neurology | 1.36 (1.23, 1.51) |
| | Pulmonary Medicine | 1.44 (1.30, 1.59) |
| | Other | 1.92 (1.79, 2.05) |
| **Modality** | **Body Group** | **Ratio of Odds Ratios** |
| **CT** | Chest | Ref |
| | Head | 0.59 (0.51, 0.67) |
| | Neck | 3.41 (3.20, 3.63) |
| | Mixed | 1.13 (1.02, 1.24) |
| **MRI** | Chest | 0.25 (0.16, 0.39) |
| | Head | 0.12 (0.04, 0.34) |
| | Neck | 0.27 (0.10, 0.75) |
| | Mixed | 0.39 (0.14, 1.06) |
| **Nuclear Medicine** | Chest | 3.07 (2.05, 4.58) |
| | Head | 0.49 (0.12, 1.99) |
| | Neck | 25.54 (9.77, 66.00) |
| | Mixed | 0.87 (0.33, 2.21) |
| **PET** | Chest | 5.90 (3.74, 9.30) |
| | Head | 0.14 (0.04, 0.56) |
| | Neck | 4.83 (1.32, 17.89) |
| | Mixed | 0.73 (0.27, 1.96) |
| **Ultrasound** | Chest | 0.07 (0.03, 0.18) |
| | Head | 0.17 (0.02, 1.60) |
| | Neck | 0.12 (0.02, 0.83) |
| | Mixed | 0.38 (0.03, 5.73) |

**Supplemental Table 9.** Thyroid cancer histology and sizes.

|  | Incidental Findings | | | P-value |
|---|---|---|---|---|
|  | No (N=106606) | Yes (N=9077) | Total (N=115683) |  |
| **Histology, n (%)** |  |  |  | 0.2547[1] |
| Anaplastic Thyroid Carcinoma | 1 (4.8%) | 0 (0.0%) | 1 (0.8%) |  |
| Follicular Thyroid Carcinoma | 0 (0.0%) | 2 (1.8%) | 2 (1.5%) |  |
| Hurthle Cell Carcinoma | 1 (4.8%) | 6 (5.5%) | 7 (5.4%) |  |
| Medullary Thyroid Carcinoma | 0 (0.0%) | 4 (3.7%) | 4 (3.1%) |  |
| Metastasis from external source | 0 (0.0%) | 1 (0.9%) | 1 (0.8%) |  |
| Papillary Thyroid Carcinoma | 19 (90.5%) | 96 (88.1%) | 115 (88.5%) |  |
| Missing | 106585 | 8968 | 115553 |  |
| **Size (mm)** |  |  |  | 0.0499[2] |
| Mean (SD) | 12.8 (7.81) | 19.7 (14.43) | 18.6 (13.82) |  |
| Missing | 106586 | 8969 | 115555 |  |

[1]Chi-Square p-value; [2]Kruskal-Wallis p-value.

**Supplemental Figure 1.** Annotation flowchart.

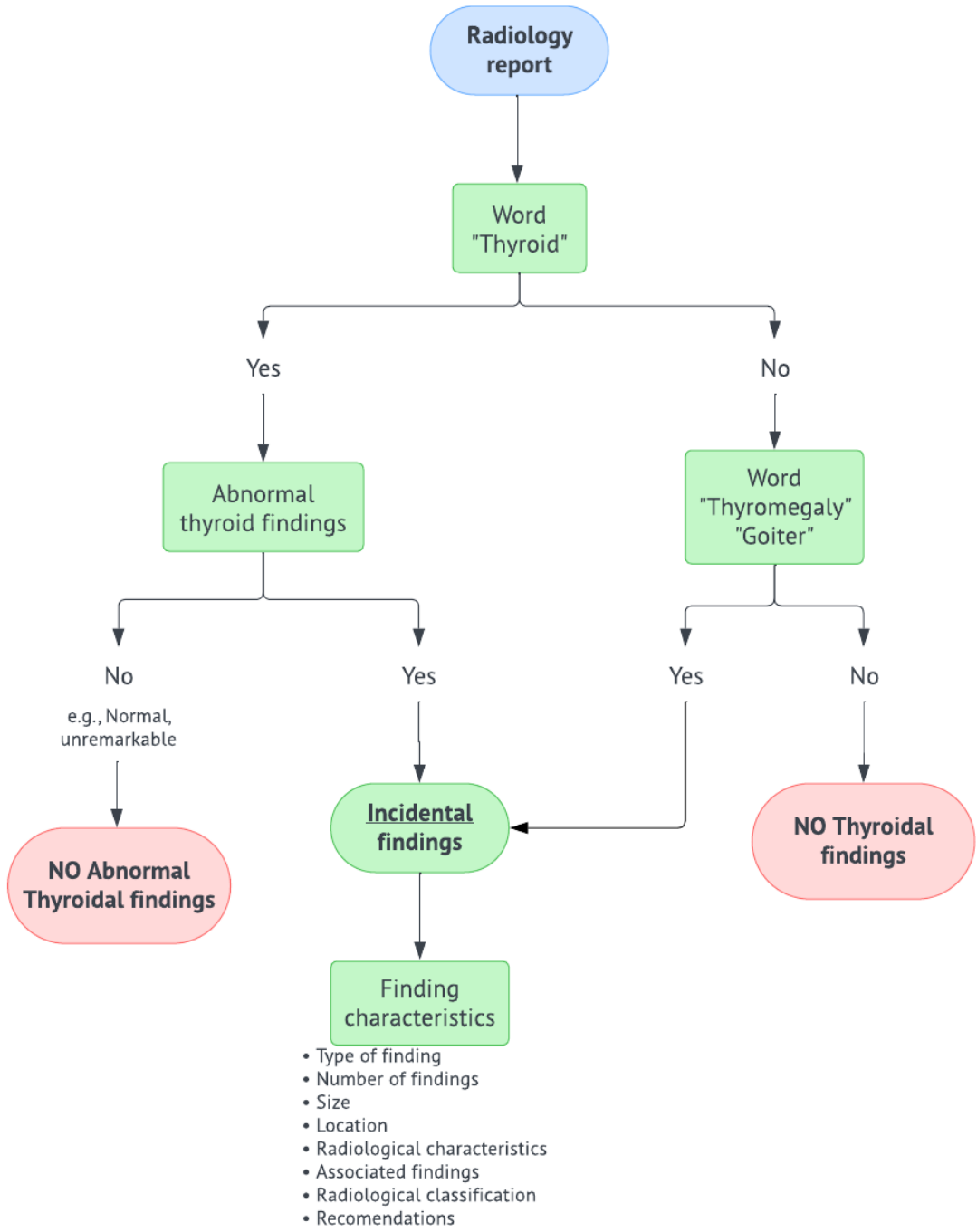

**Supplemental Figure 2.** An overview of the incidental nodular thyroid reports identification pipeline.

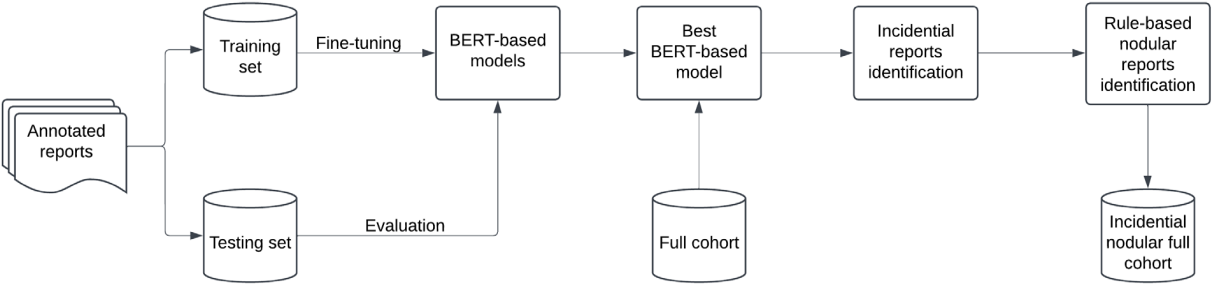

**Supplemental Figure 3.** An overview of the target entities recognition pipeline using incidental nodular thyroid reports.

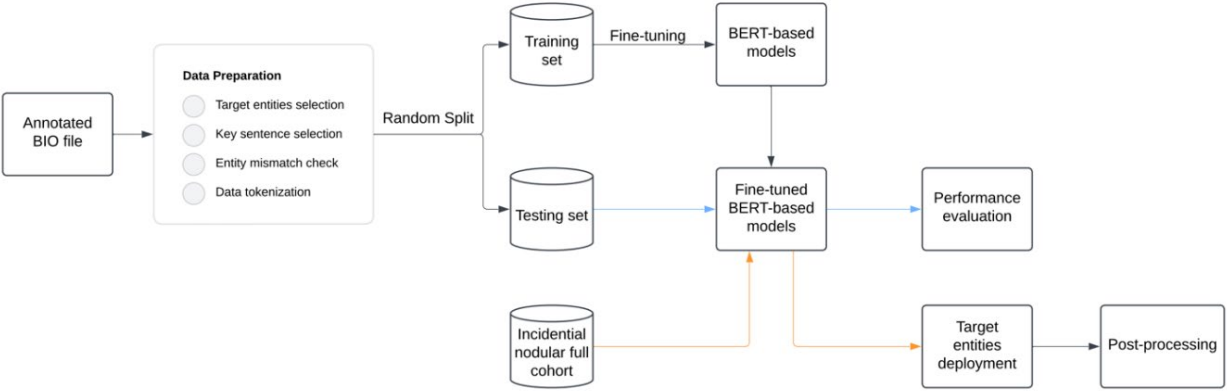

**Supplemental Figure 4.** Interactions between body location and imaging modality.

MRI: Magnetic Resonance Imaging; PET: Positron Emission Tomography; CT: Computed Tomography.

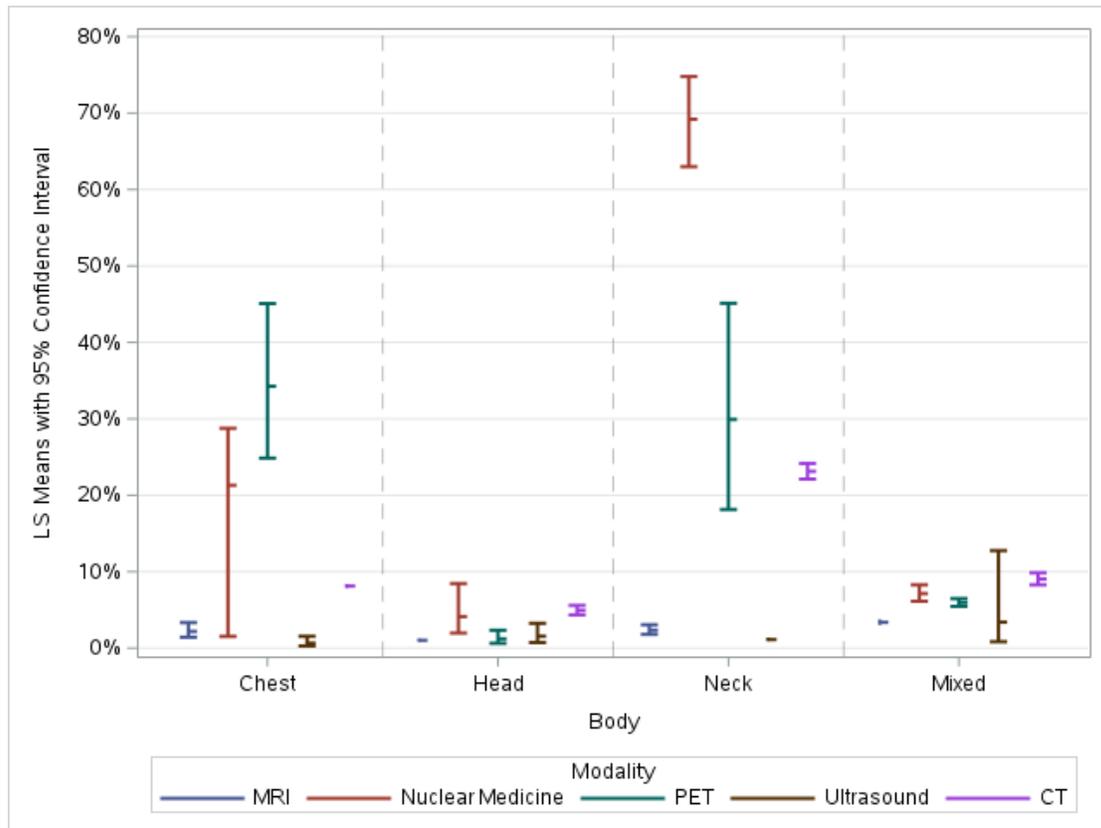